\documentclass[10pt,conference,a4paper]{IEEEtran}
\usepackage{times}
\usepackage{epsfig}
\usepackage{graphicx}
\usepackage{amsmath}
\usepackage{amssymb}
\usepackage[caption=false,font=footnotesize,labelfont=sf,textfont=sf]{subfig}
\usepackage{mathrsfs}
\usepackage{mathtools}
\usepackage{booktabs}
\usepackage{adjustbox}
\usepackage{multirow}
\usepackage[colorlinks]{hyperref}
\usepackage{array}
\usepackage{mynotation}
\usepackage{color}
\usepackage[nocompress]{cite}
\begin{document}

%%%%%%%%% TITLE
\title{Deep Motion Features for Visual Tracking}

\author{Susanna Gladh, Martin Danelljan, Fahad Shahbaz Khan, Michael Felsberg \\
	\small Computer Vision Laboratory, Department of Electrical Engineering, Link\"oping University, Sweden
	}

\maketitle
%\thispagestyle{empty}

%%%%%%%%% ABSTRACT
\begin{abstract}
   Robust visual tracking is a challenging computer vision problem, with many real-world applications. Most existing approaches employ hand-crafted appearance features, such as HOG or Color Names. Recently, deep RGB features extracted from convolutional neural networks have been successfully applied for tracking. Despite their success, these features only capture appearance information. On the other hand, motion cues provide discriminative and complementary information that can improve tracking performance. Contrary to visual tracking, deep motion features have been successfully applied for action recognition and video classification tasks. Typically, the motion features are learned by training a CNN on optical flow images extracted from large amounts of labeled videos.

   This paper presents an investigation of the impact of deep motion features in a tracking-by-detection framework. We further show that hand-crafted, deep RGB, and deep motion features contain complementary information. To the best of our knowledge, we are the first to propose fusing appearance information with deep motion features for visual tracking.
   Comprehensive experiments clearly suggest that our fusion approach with deep motion features outperforms standard methods relying on appearance information alone.

\end{abstract}

%%%%%%%%% BODY TEXT
\section{Introduction}

Generic visual tracking is the problem of estimating the trajectory of a target in a sequence of images. It is challenging since only the initial position of the target is known. The various applications of generic tracking range from surveillance to robotics. Most state-of-the-art approaches follow the tracking-by-detection paradigm, where a classifier or regressor is discriminatively trained to differentiate the target from the background.
Recently, Discriminative Correlation Filter (DCF) based methods \cite{MOSSE2010,DanelljanBMVC14,DanelljanICCV2015,Henriques14,HCF_ICCV15} have shown excellent performance for visual tracking. These approaches efficiently train a correlation filter to estimate the classification confidences of the target. This is performed by considering all cyclic shifts of the training samples and exploiting the FFT at the training and detection steps. In this work, we base our approach on the DCF framework.

DCF based trackers typically employ hand-crafted appearance features, such as HOG \cite{DanelljanBMVC14,Henriques14}, Color Names \cite{DanelljanCVPR14}, or combinations of these features \cite{Li2014}. 
Recently, deep Convolutional Neural Networks (CNNs) has been successfully applied for tracking \cite{DanelljanECCV2016,DanelljanVOT2015,HCF_ICCV15, Nam2016CVPR}. A CNN applies a sequence of convolution, normalization, and pooling operations on the input RGB patch. The parameters of the network are trained using large amounts of labeled images, such as the ImageNet dataset \cite{ILSVRC15}. Deep convolutional features from pre-trained networks have been shown to be generic \cite{RazavianASC14}, and therefore also applicable for visual tracking.
\begin{figure}[!t]
	\centering
	\newcommand{\wid}{0.16\textwidth}
	\newcommand{\name}{figures/cle_bs}
	\includegraphics[trim={0 60 0 40},clip,width=\wid]{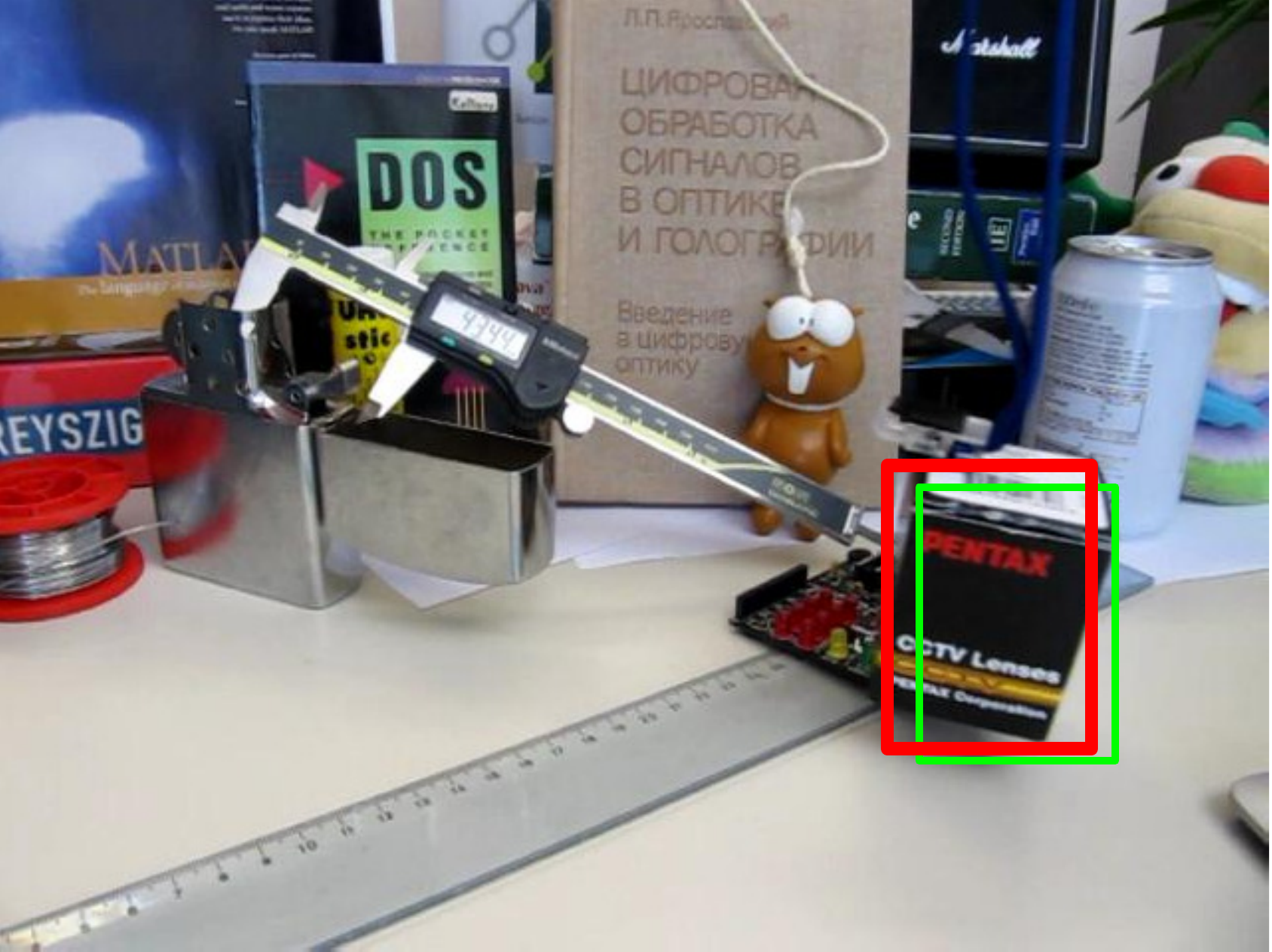}%
	\includegraphics[trim={0 60 0 40},clip,width=\wid]{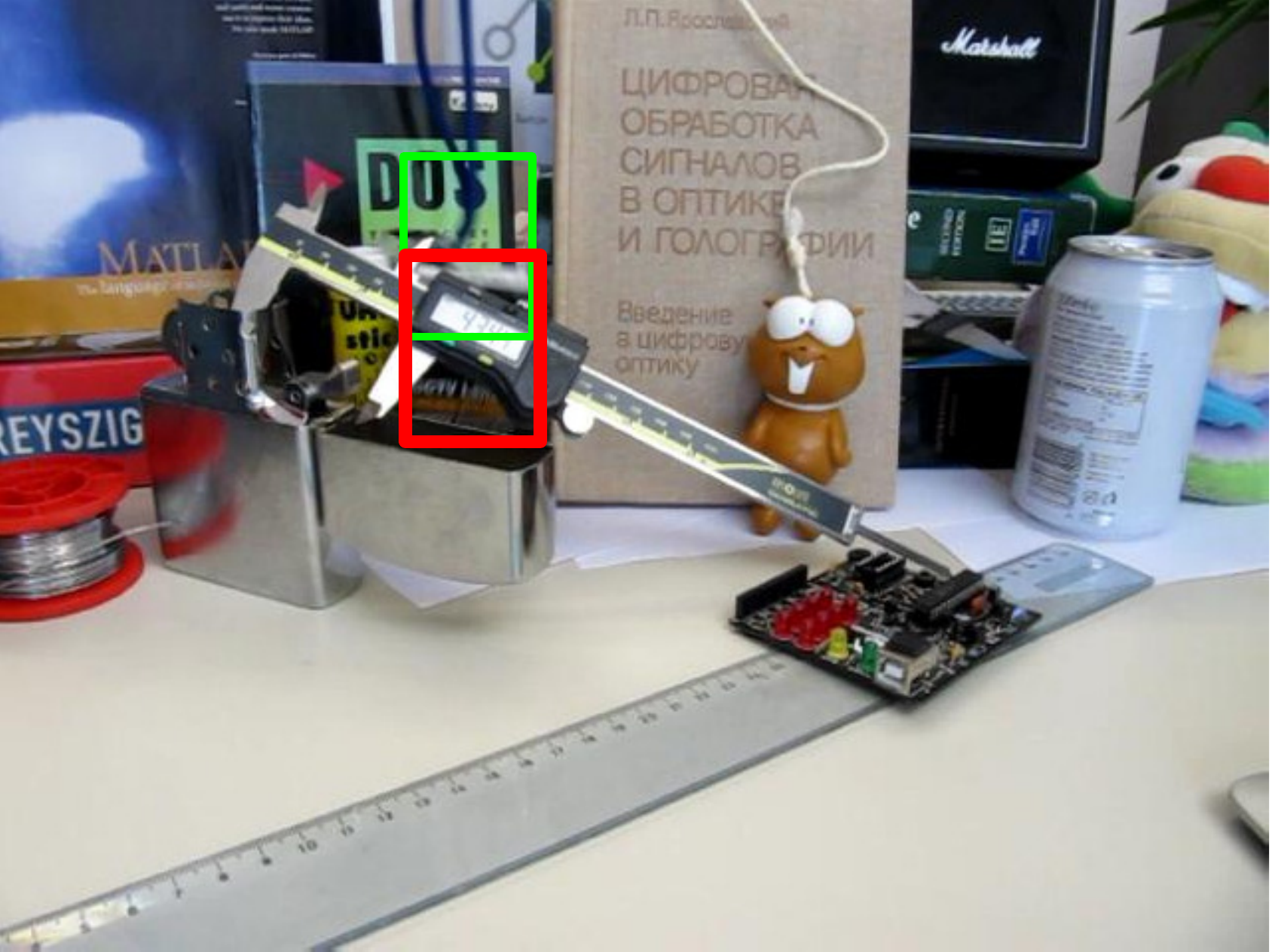}%
	\includegraphics[trim={0 60 0 40},clip,width=\wid]{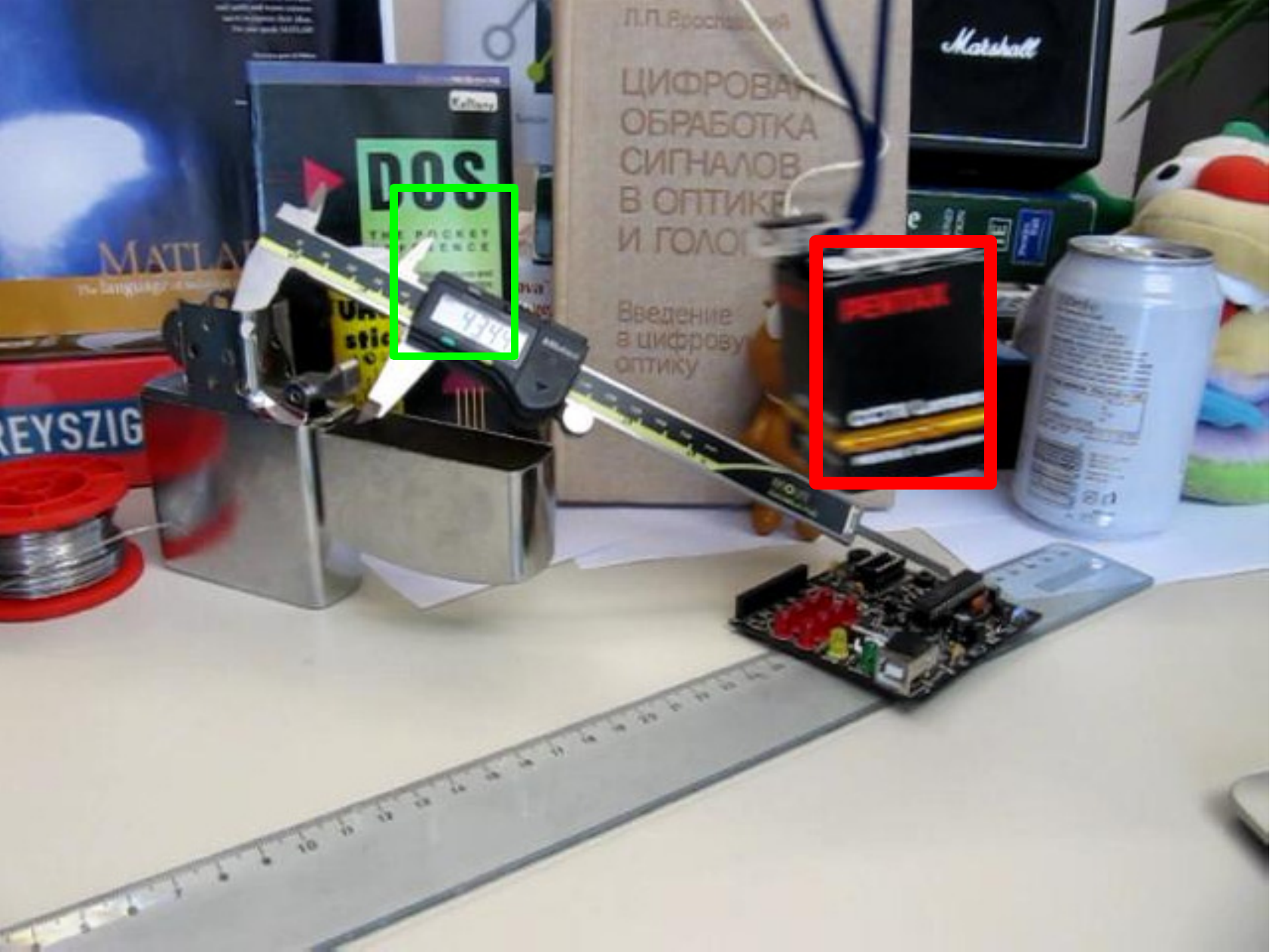}
	\includegraphics[trim={0 35 0 30},clip,width=\wid]{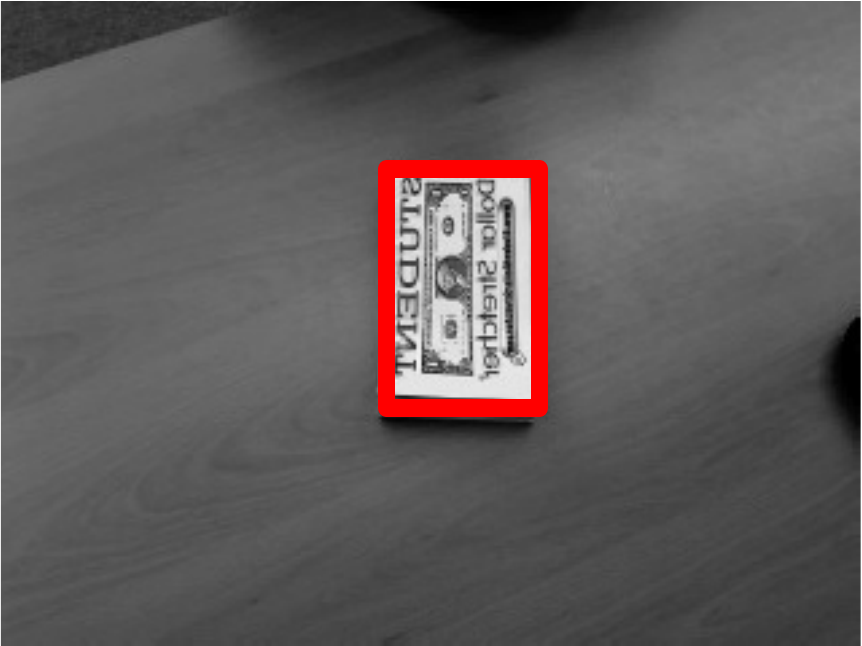}%
	\includegraphics[trim={0 35 0 30},clip,width=\wid]{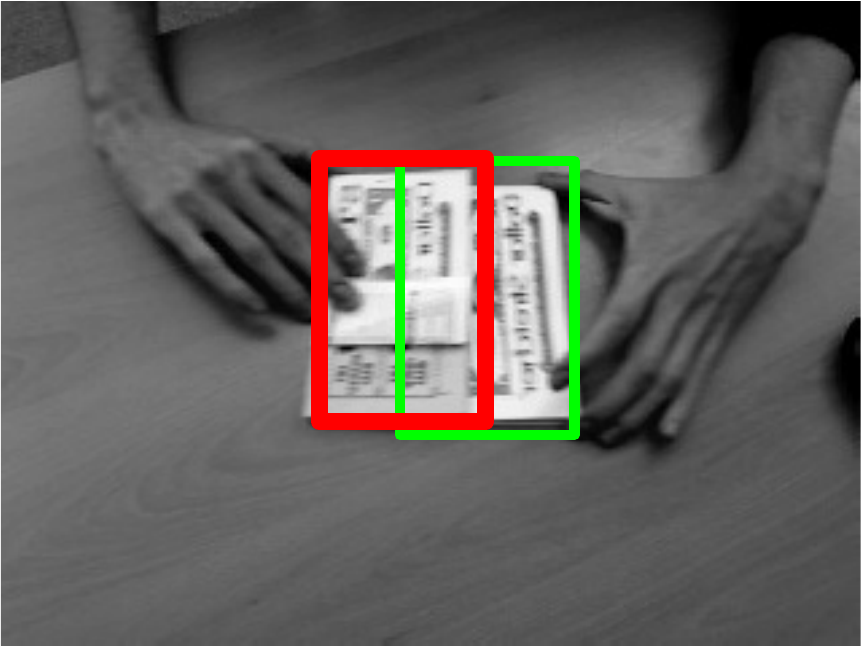}%
	\includegraphics[trim={0 35 0 30},clip,width=\wid]{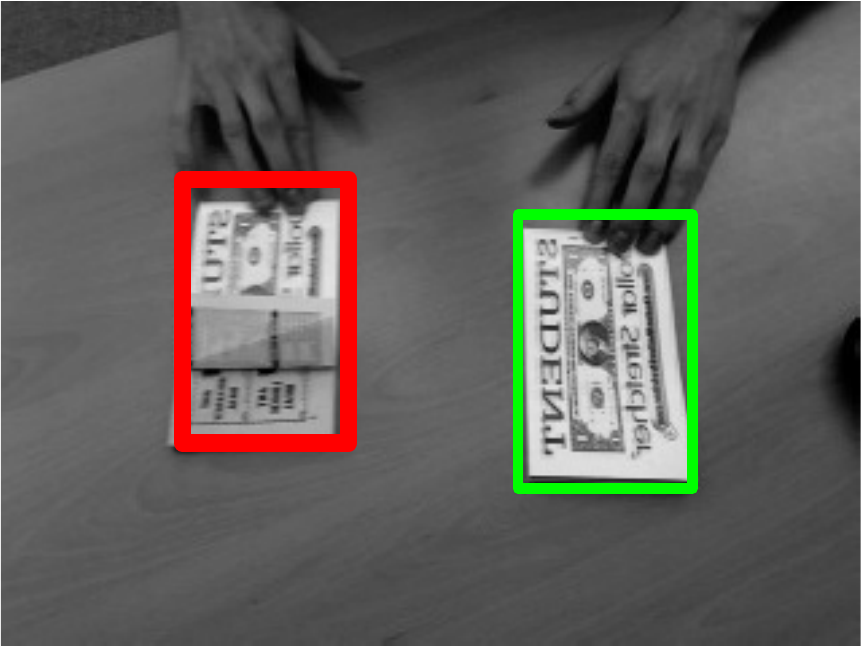}
	\includegraphics[trim={0 20 0 20},clip,width=\wid]{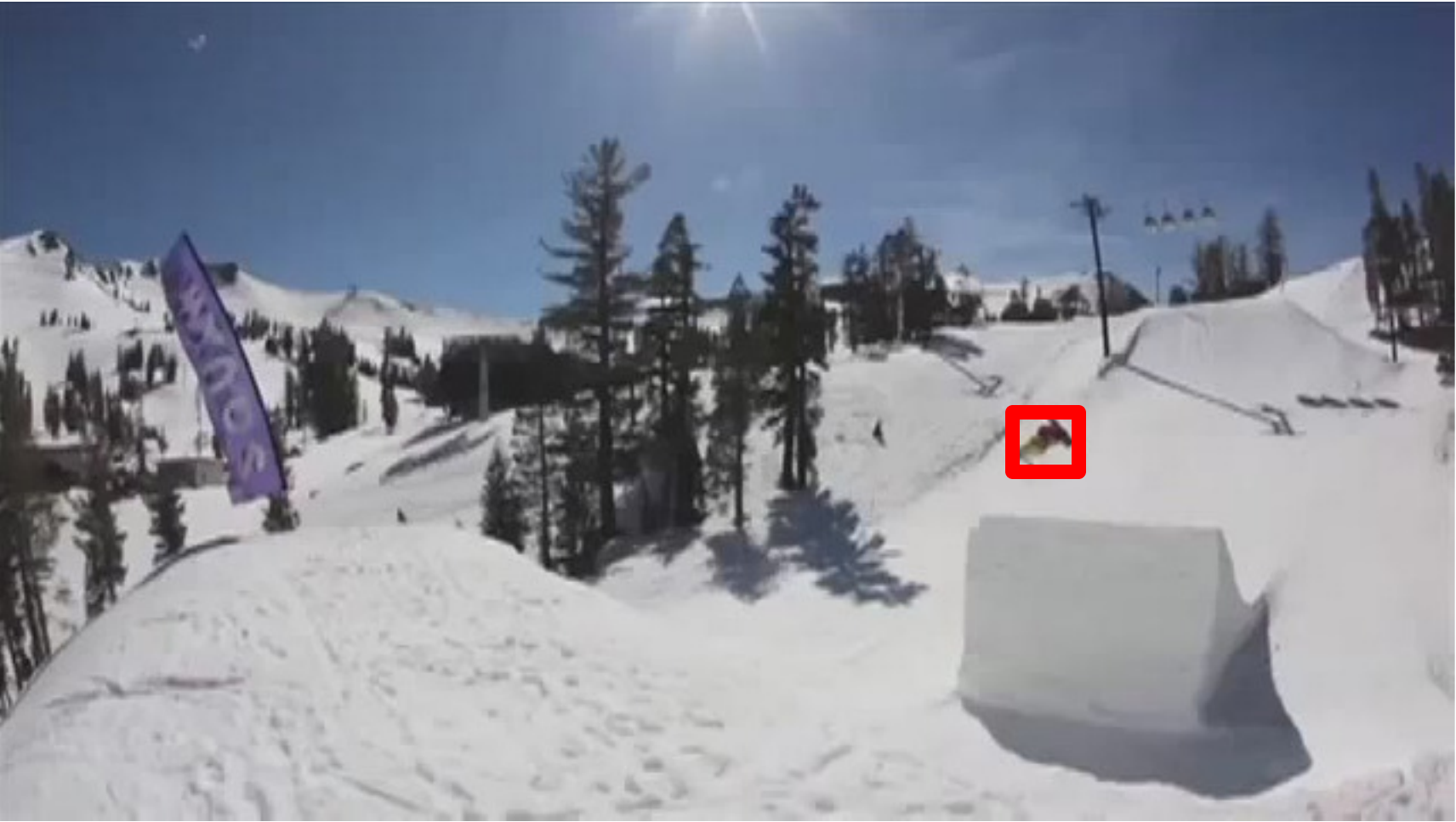}%
	\includegraphics[trim={0 20 0 20},clip,width=\wid]{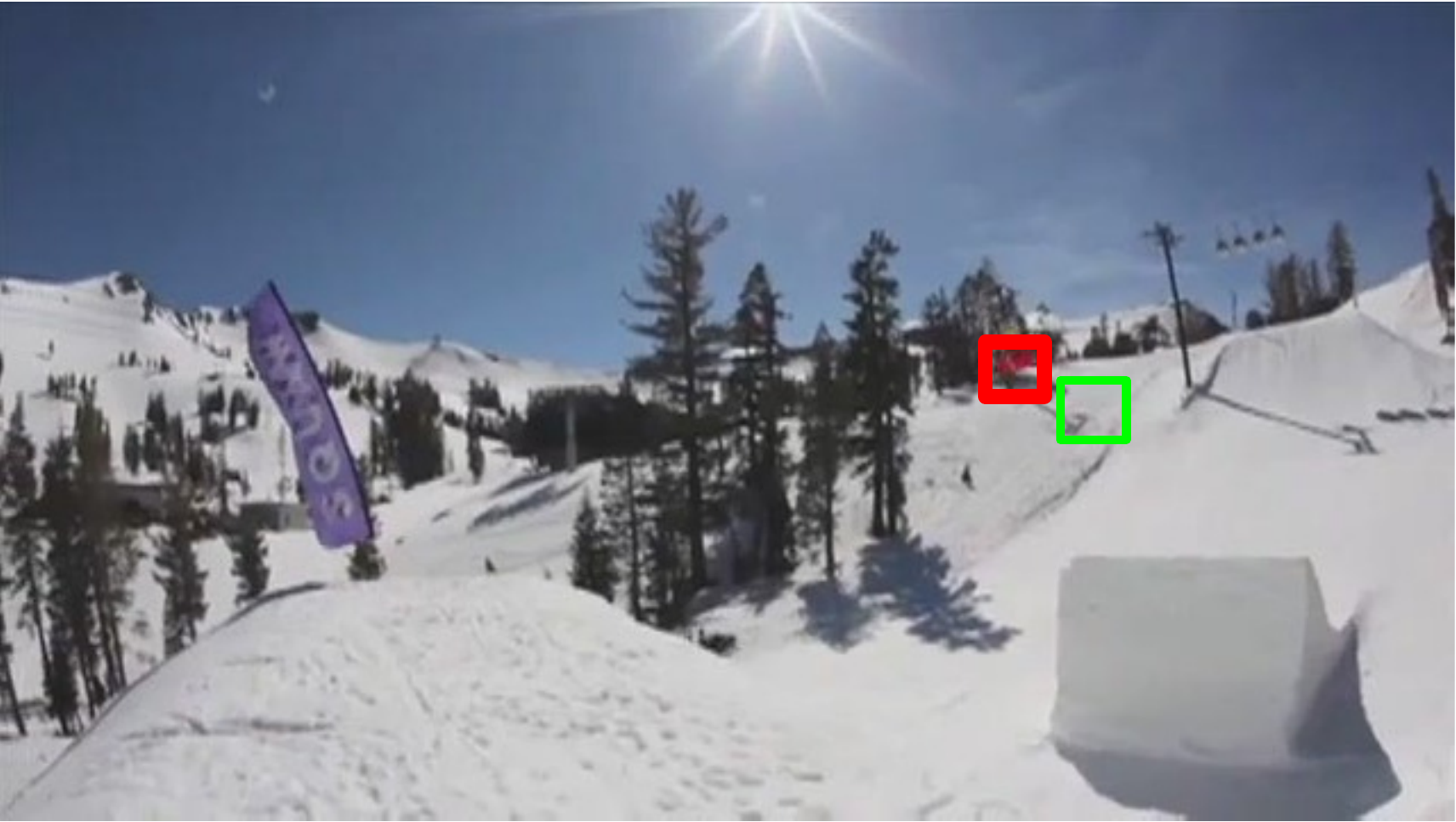}%
	\includegraphics[trim={0 20 0 20},clip,width=\wid]{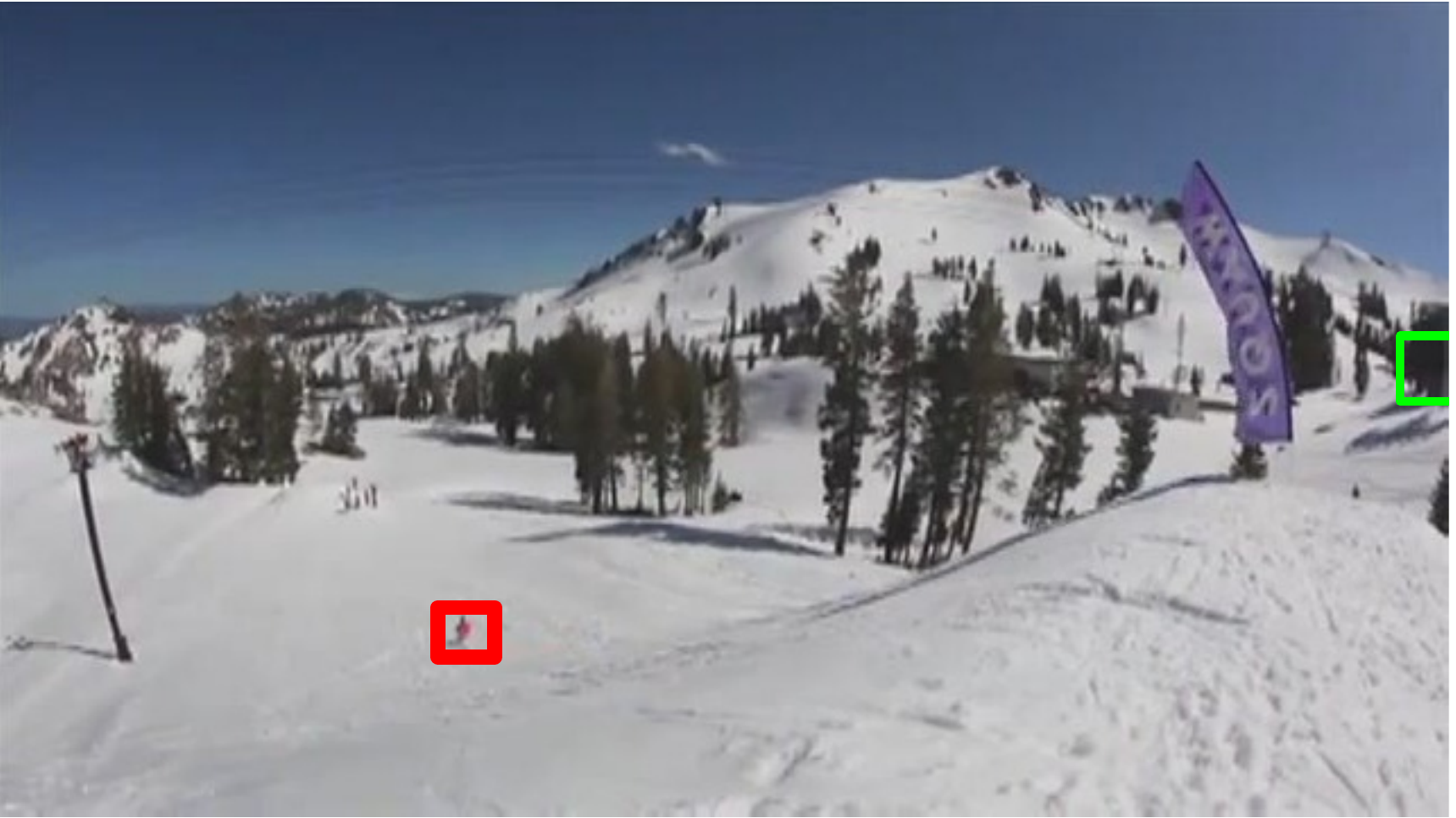}\vspace{-1mm}
	\includegraphics*[trim=5 5 5 5,width=\columnwidth]{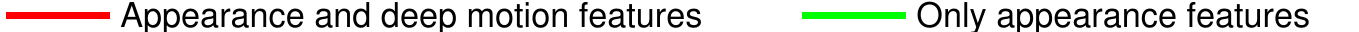}\vspace{-2mm}
	\caption{A comparison of using combined appearance information (hand-crafted HOG and deep RGB features) (in green) and our fusion of appearance and deep motion features (in red). Tracking results are shown for three example sequences: \emph{Box}, \emph{Coupon} and \emph{Skiing}. Our fusion approach (red), using deep motion features, achieves superior results in these scenarios where appearance alone is insufficient.  }
	\label{fig:intro}
	\vspace{-5mm}
\end{figure}

Besides deep RGB features, deep motion features have been successfully employed for action recognition \cite{actiontubes,SimonyanZ14}. These motion features are constructed by learning a CNN that takes optical flow images as input. The network is trained using flow data extracted from large amounts of labeled videos, such as the UCF101 dataset \cite{UCF101}. Unlike deep RGB networks, these deep flow networks capture high-level information about the motion in the scene. To the best of our knowledge, deep motion features are yet to be investigated for the problem of visual tracking.

Tracking methods solely based on appearance information struggle in scenarios with, for example, out-of-plane rotations (figure~\ref{fig:intro} first row), background distractors with similar appearance (figure~\ref{fig:intro} second row), and distant or small objects (figure~\ref{fig:intro} third row). In these cases, high-level motion cues provide rich complementary information that can disambiguate the target. While appearance features only encode static information from a single frame, deep motion features integrate information from the consecutive \emph{pair} of frames used for estimating the optical flow. Motion features can therefore capture the dynamic nature of a scene that is complementary to appearance features. This motivates us to investigate the fusion of standard appearance features with deep motion features for visual tracking.

\noindent\textbf{Contributions:}
In this paper, we investigate the impact of deep motion features for visual tracking. We use a deep optical flow network that was pre-trained for action recognition. Our approach does not require any additional labeled data for training the network. We investigate fusing hand-crafted and deep appearance features with deep motion features in a state-of-the-art DCF-based tracking-by-detection framework \cite{DanelljanICCV2015}. To show the impact of motion features, we further evaluate the fusion of different feature combinations.

Extensive experiments are performed on the OTB-2015 \cite{OTB2015}, Temple-Color \cite{TempleColor}, and VOT-2015 \cite{VOT2015} datasets. On OTB-2015, our fusion of appearance and deep motion features significantly improves the baseline method, employing only appearance features, by $3.4 \%$ in mean overlap precision. Our fusion approach is further shown to advance the state-of-the-art performance with an absolute gain of $6.8\%$ and $5.8\%$ in mean overlap precision on OTB-2015 and Temple-Color respectively. Figure~\ref{fig:intro} shows a comparison of our fusion with deep motion features with the baseline method (employing only appearance features).

\section{Related Work}

Discriminative tracking methods \cite{Torr11b,MEEM2014,DanelljanICCV2015,HCF_ICCV15} typically train a classifier or regressor for the task of differentiating the target from the background. These methods are often also termed tracking-by-detection approaches since they apply a discriminatively trained detector. Learning is performed online by extracting and labeling samples from the video frames. The training samples are often represented by e.g. raw image patches \cite{MOSSE2010,Henriques12}, image histograms and Haar features \cite{Torr11b}, color \cite{DanelljanCVPR14,MEEM2014}, or shape features \cite{DanelljanBMVC14,Henriques14}. Lebeda \textit{et al.} \cite{Lebeda13} use edge-points and estimate the transformation between each frame to successfully perform tracking on texture-less objects. C\^{e}hovin \textit{et al.} \cite{CehovinLGT} employ a couple-layer visual model, which combine local and global target appearance, to perform robust tracking of objects that have rapid appearance changes. 

Among tracking-by-detection approaches, the Discriminative Correlation Filter (DCF) based trackers have recently shown excellent performance on standard tracking benchmarks \cite{Wu13,VOT2015}. The key for their success is the ability to efficiently utilize limited data by including all shifts of local training samples in the learning. DCF-based methods train a least-squares regressor for predicting the target confidence scores by utilizing the properties of circular correlation and the fast Fourier transform (FFT). The MOSSE tracker \cite{MOSSE2010} first considered training a single-channel correlation filter based on grayscale image samples of target and background appearance. A remarkable improvement is achieved by extending the MOSSE filter to multi-channel features. This can be performed by either optimizing the exact filter for offline learning applications \cite{galoogahiICCV13} or using approximative update schemes for online tracking \cite{DanelljanBMVC14,DanelljanCVPR14,Henriques14}.

Despite their success, the DCF approaches are affected by the periodic assumption of the training samples, leading to negative boundary effects and a restricted training and search region size. This problem has recently been addressed in \cite{DanelljanICCV2015} by adding a spatial regularization term. While the original SRDCF employs HOG features, the DeepSRDCF \cite{DanelljanVOT2015} investigates the use of convolutional features from a deep RGB network in the SRDCF tracker. In this work, we also base our tracking framework on the SRDCF.

Other than the deep RGB appearance features, recent works \cite{GaloogahiCVPR2015,CheronICCV2015,SimonyanZ14} have investigated the use of deep motion features for action recognition. Generally, optical flow is computed for each consecutive pair of frames. The resulting optical flow is aggregated in the x-, y- direction and the flow magnitude to construct a three channel image. A CNN is then trained using these flow patches as input. Simonyan and Zisserman \cite{SimonyanZ14} propose a two-stream ConvNet architecture to integrate spatial and temporal networks. The network is trained on multi-frame dense optical flow and multi-task learning is employed to increase the amount of training samples. Gkioxari and Malik \cite{GaloogahiCVPR2015} propose to use deep static and kinematic cues for action localization in videos. The work of \cite{CheronICCV2015} propose to combine pose-normalized deep appearance and motion features for action recognition. Unlike action recognition, existing tracking methods \cite{DanelljanVOT2015,HCF_ICCV15} only utilize appearance based deep RGB features. In this work, we propose to combine appearance cues with deep motion information for visual tracking.

\section{Baseline Tracker} \label{sec:BT}
As a baseline tracker, we employ the SRDCF \cite{DanelljanICCV2015} framework, which has recently been successfully used for integrating single-layer deep features \cite{DanelljanVOT2015}. The standard DCF trackers exploit the periodic assumption of the local feature map to perform efficient training and detection using the FFT. However, this introduces unwanted boundary effects and restricts the size of the image region used for training the model and searching for the target. In the SRDCF, these shortcomings are addressed by introducing a spatial regularization term in the learning formulation. This enables training to be performed on larger image regions, leading to a more discriminative model.

In the SRDCF framework, a convolution filter is discriminatively learned based on training samples $\{(x_k, y_k)\}_{k=1}^{t}$. Here, $x_k$ is a $d$-dimensional feature map with a spatial size $M \times N$. We denote feature channel $l$ of $x_k$ by $x_k^l$. Typically, $x_k$ is extracted from an image region containing both the target and large amounts of background information. The label $y_k$ consists of the desired $M \times N$ confidence score function at the spatial region corresponding to the sample $x_k$. That is, $y_k(m,n) \in \reals$ is the desired classification confidence at location $(m,n)$ in the feature map $x_k$. We use a Gaussian function centered at the target location in $x_k$ to determine the desired scores $y_k$. In the SRDCF formulation, the aim is to train a multi-channel convolution filter $f$ consisting of one $M \times N$ filter $f^l$ per feature dimension $l$. The target confidence scores for an $M \times N$ feature map $x$ are computed as $S_f(x) = \sum_{l=1}^d x^l \conv f^l $.
Here, $\conv$ denotes circular convolution.

To learn the filter $f$, the SRDCF formulation minimizes the squared error between the confidence scores $S_f(x_k)$ and the corresponding desired scores $y_k$,
\begin{equation}
\label{eq:ours_cost_spatial}
\varepsilon(f) = \sum_{k=1}^t \alpha_k \big\| S_f(x_k) - y_k \big\|^2 + \sum_{l=1}^d \big\| w \pmult f^l \big\|^2 .
\end{equation}
The weights $\alpha_k$ determine the impact of each training sample and $\pmult$ denotes point-wise multiplication. The SRDCF employs a spatial regularization term determined by the penalty weight function $w$. The filter is trained by minimizing the least squares loss \eqref{eq:ours_cost_spatial} in the Fourier domain using iterative sparse solvers. We refer to \cite{DanelljanICCV2015} for more details about the training procedure. Target detection is explained further in section \ref{sec:our_tracking_framework}.

%VISUAL FEATURES
\section{Visual features}
The visual feature representation is a core component of a tracking framework. In this work, we investigate the use of a combination of hand-crafted features, deep appearance features and deep motion features for tracking.

\subsection{Hand-crafted features}
Hand-crafted features are typically used to capture low-level information, such as shape, color or texture. The Histograms of Oriented Gradients (HOG) is popularly employed for both visual tracking \cite{DanelljanBMVC14,DanelljanICCV2015,Henriques14} and object detection \cite{pedro10}. It mainly captures shape information by calculating histograms of gradient directions in a spatial grid of cells. The histogram for each cell is then normalized with respect to neighboring cells to add invariance.

Other than shape features, various color-based feature representations have been commonly employed for tracking. For instance, the use of simple color transformations \cite{Oron12b,MEEM2014} or color histograms \cite{DanelljanSCIA2015}. Recently, the Color Names (CN) descriptor have been popularly employed for tracking due to its discriminative power and compactness \cite{DanelljanCVPR14}. The CN descriptor applies a pre-learned mapping from RGB to the probabilities of 11 linguistic color names.

\subsection{Deep features}
Features extracted by a trained Convolutional Neural Network (CNN) are known as deep features. The CNN consists of a number of computational layers that perform convolution, local normalization, and pooling operations on the input image patch. The final layers are usually fully connected (FC) and include an output classification layer. CNNs are typically trained in a supervised manner on large datasets of labeled images, such as ImageNet \cite{ILSVRC15}.

Feature representations learned by CNNs trained for a particular vision problem (e.g. image classification) have been shown to be generic and can be applied for a variety other vision tasks. For this purpose, most works apply the activations from the FC layer \cite{SimonyanICLR2015,OquabCVPR2014}. Recently, activations from the convolutional layers have shown improved results for image classification \cite{VedaldiCVPR2015,HengelCVPR2015}. The deep convolutional features are discriminative and posses high-level visual information, while preserving spatial structure. Convolutional activations at a specific layer from a multi-channel feature map, which can be directly integrated into the SRDCF framework. Shallow-layer activations encode low-level features at a high spatial resolution, while deep layers contain high-level information at a coarse resolution.

\begin{figure}[!t]
	\centering
	\includegraphics[width=\linewidth]{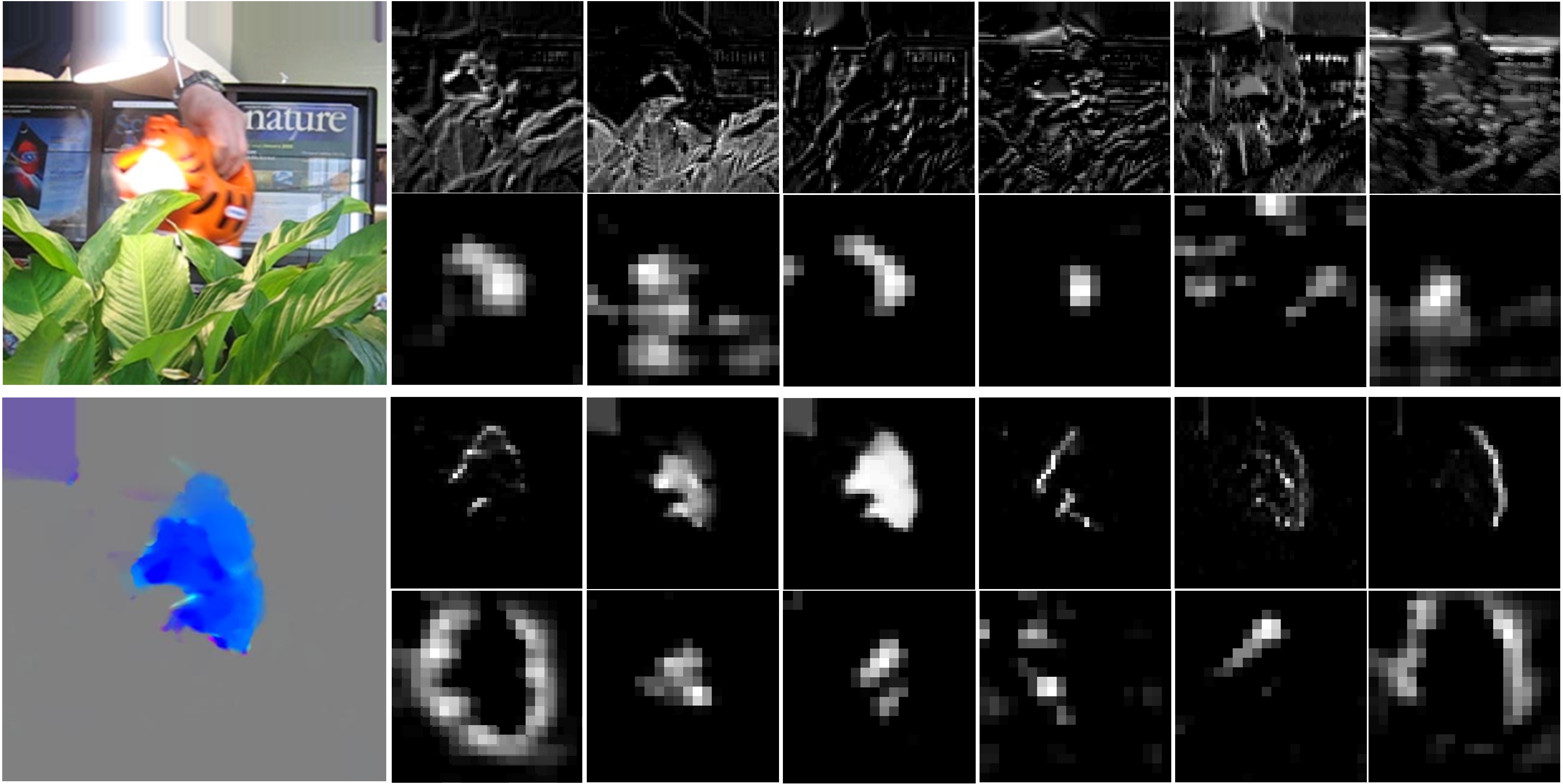}\vspace{-2mm}
	\caption{Visualization of the features with highest energy from a shallow and deep convolutional layer in the appearance (top row) and motion network (bottom row). Appearance features are extracted from the raw RGB image (top left) from Tiger2, and motion features from the corresponding optical flow image (bottom left). In both cases, we show shallow and deep activations in the corresponding first and second sub-rows respectively.}\vspace{-3mm}
	\label{fig:features}
\end{figure}
\noindent\textbf{Deep RGB Features:}
For the RGB images we use the imagenet-vgg-verydeep-16 network \cite{SimonyanICLR2015}, with the MatConvNet library \cite{matconvnet}. This network contains 13 convolutional layers. We investigate using both a shallow and a deep convolutional layer. For the shallow RGB layer, we use the activations from the fourth convolutional layer, after the Rectified Linear Unit (ReLU) operation. It consists of 128 feature channels and has a spatial stride of 2 pixels compared to the input image patch. For the deep layer of the RGB network, we use activations at the last convolutional layer, again after the ReLU-operation. This layer consists of a 512-dimensional feature map with a spatial stride of 16 pixels. Figure~\ref{fig:features} shows example activations from the shallow layer (first row) and deep layer (second row) of the RGB network.

\noindent\textbf{Deep Motion Features:}
The motion features are extracted using the approach described by \cite{CheronICCV2015}. We start by calculating the optical flow for each frame, together with the previous frame, according to \cite{bro04a}. The motion in the x- and y-directions forms a 3-channel image together with the flow magnitude. The values are adjusted to the interval $[0, 255]$.
For our experiments, we use the pre-trained optical flow network provided by \cite{actiontubes}. It is pre-trained on the UCF101 dataset \cite{UCF101} for action recognition and contains five convolutional layers. For the motion network, we only use the activations from the deepest convolutional layer. Similar to the RGB network, we extract the activations after the ReLU-operation. The resulting feature map consists of 384 channels at a spatial stride of 16 pixel compared to the input. An example optical flow image is displayed in figure~\ref{fig:features}, along with corresponding shallow (third row) and deep (fourth row) activations from the motion network.
\begin{table*}[!t]
	\centering
	\caption{Comparison of different combinations of hand-crafted (HOG), deep RGB and deep motion features on the OTB-2015 dataset. Results are reported in terms of mean overlap precision (OP) and area-under-the-curve (AUC) in percent. The two best results are displayed in red and blue font respectively. Shallow and deep layers of the RGB network are denoted RGB(s) and RGB(d). For each combination of appearance features, we also report the results obtained when including deep motion features. We omit the result of shallow motion features, since we did not observe a gain in performance when including them. The fusion with deep motion features significantly improves the performance for all combinations.
		\vspace{-2mm}
	}
	\resizebox{0.9\textwidth}{!}{\begin{tabular}{l l c c c c c c c }
\toprule
& & HOG&RGB(s)&RGB(d)&RGB(s+d)&HOG+RGB(s)&HOG+RGB(d)&HOG+RGB(s+d) \\
\midrule
\multirow{2}{*}{\textbf{Mean OP (\%)}} & Without deep motion features &74.5&74.1&56.3&78&74.9&80.7&79.1\\
 & With deep motion features &81.3&81&58.9&81.1&79.5&\textbf{\textcolor{red}{84.1}}&\textit{\textcolor{blue}{82.2}}\\
\midrule
\multirow{2}{*}{\textbf{Mean AUC (\%)}} & Without deep motion features &61.1&62.8&48.5&65.1&62.6&65.2&65.3\\
 & With deep motion features &65.7&66.4&49.7&\textit{\textcolor{blue}{66.7}}&65.6&\textbf{\textcolor{red}{67.4}}&66.4\\
\bottomrule
\end{tabular}
}
	\vspace{-1mm}
	\label{tab:baseline}
	\vspace{0mm}
\end{table*}
\section{Our Tracking Framework} \label{sec:our_tracking_framework}

Here, we describe our tracking framework where we investigate the fusion of hand-crafted and deep appearance features with deep motion features. Our framework is based on learning an independent SRDCF model for each feature map. That is, we learn one filter $f_j$ for each feature type $j$. In a frame $k$, we extract new training samples $x_{j,k}$ for each feature type $j$ from the same image region centered at the estimated target location. We use a quadratic training region with an area equal to $5^2$ times the area of the target box. For example, in our final tracking approach we combine three different feature maps: HOG $x_{1,k}$, the deep RGB layer $x_{2,k}$ and the deep motion layer $x_{3,k}$ (see section~\ref{sec:baselinecomparison}). The fused feature maps have different dimensionalities $d_j$ and spatial resolutions, leading to a different spatial sample size $M_j \times N_j$ for each feature $j$. The label function $y_{j,k}$ for feature $j$ is set to an $M_j \times N_j$ sampled Gaussian function with its maximum centered at the estimated target location.

To train the filters, we minimize the SRDCF objective \eqref{eq:ours_cost_spatial} for each feature type $j$ independently. This is performed similarly to \cite{DanelljanICCV2015} by first transforming \eqref{eq:ours_cost_spatial} to the Fourier domain using Parseval's formula and then applying an iterative solver. We also use exponentially decreasing sample weights $\alpha_k$ \cite{MOSSE2010,DanelljanBMVC14,DanelljanICCV2015} with a learning rate of $0.01$ and construct the penalty function $w$ as in \cite{DanelljanICCV2015}.

To detect the target in a new frame, we first extract feature maps $z_j$ centered at the estimated target location in the previous frame. This is performed using the same procedure as for training samples. The learned filters $f_j$ from the previous frame can then be applied to each feature map $z_j$ individually. However, the target confidence scores $S_{f_j}(z_j)$ is of size $M_j \times N_j$ and therefore have a different spatial resolution for each feature type $j$. To fuse the confidence scores obtained from each filter $f_j$, we first interpolate the scores from each filter to a pixel-dense grid. We then fuse the scores by computing the average confidence value at each pixel location. For efficiency, we use the Fourier interpolation method employing complex exponential basis functions. Since the filters are optimized in the Fourier domain, we directly have the DFT coefficients $\hat{f}_j$ of each filter. Using the DFT convolution property, the DFT coefficients of the confidences are obtained as $\widehat{S_{f_j}(z_j)} = \sum_{l=1}^{d_j} \hat{z}^l_j \pmult \hat{f}^l_j$.

\begin{table*}[!t]
	\centering
	\caption{Comparison with state-of-the-art trackers using mean OP ($\%$) on the OTB-2015 and Temple-Color datasets. The two best results are show in red and blue font respectively. Our approach significantly improves the state-of-the-art DeepSRDCF tracker by $6.8\%$ and $5.8\%$ on OTB-2015 and Temple-Color datasets respectively. }
	\vspace{-2mm}
	\resizebox{0.9\textwidth}{!}{\begin{tabular}{l@{~}c@{~~}c@{~~}c@{~~}c@{~~}c@{~~}c@{~~}c@{~~}c@{~~}c@{~~}c@{~~}c@{~~}c@{~~}c@{~~}c}
\toprule
&Struck&CFLB&ACT&KCF&DSST&SAMF&DAT&MEEM&LCT&HCF&SRDCF&SRDCFdecon&DeepSRDCF&\textbf{Ours}\\\midrule
OTB-2015&52.9&44.9&49.6&54.9&60.6&64.7&36.4&63.4&70.1&65.5&72.9&76.5&\textit{\textcolor{blue}{77.3}}&\textbf{\textcolor{red}{84.1}}\\
Temple-Color&40.9&37.8&42.1&46.5&47.5&56.1&48.2&62.2&52.8&58.2&62.2&65&\textit{\textcolor{blue}{65.4}}&\textbf{\textcolor{red}{71.2}}\\\bottomrule
\end{tabular}
}
	\vspace{0mm}
	\label{tab:otb_tpl}
	\vspace{-3mm}
\end{table*}

The Fourier interpolation is implemented by first zero-padding the DFT coefficients to the desired resolution and then performing inverse DFT. Formally, we define the padding operator $\mathcal{P}_{R\times S}$ that pads the DFT to the size $R \times S$ by adding zeros at the high frequencies. We denote the inverse DFT operator by $\ftinv$ and let $J$ denote the number of feature maps to be fused. The fused confidence scores $s$ are computed as,
\begin{align}
	\label{eq:fusion}
	s &= \frac{1}{J} \sum_{j=1}^{J} \ftinv \bigg\{ \mathcal{P}_{R\times S} \left\{\widehat{S_{f_j}(z_j)}\right\} \bigg\} 
\end{align}
We obtain pixel-dense confidence scores \eqref{eq:fusion} of the target by setting $R \times S$ to be the size (in pixels) of the image region used for extracting the feature maps $x_{j,k}$. The new target location is then estimated by maximizing the scores $s(m,n)$ over the pixel locations $(m,n)$. To also estimate the target size, we apply the filters at five scales with a relative scale factor of $1.02$, similar to \cite{DanelljanICCV2015,Li2014}.
 
\section{Experiments} \label{sec:experiments}
We validate our tracking framework by performing comprehensive experiments on three challenging benchmark datasets: OTB-2015 \cite{OTB2015} with 100 videos, Temple-Color \cite{TempleColor} with 128 videos, and VOT2015 \cite{VOT2015} with 60 videos.

\subsection{Baseline Comparison} \label{sec:baselinecomparison}
We investigate the impact of deep motion features by evaluating different combinations of appearance and motion representations on the OTB-2015 dataset. Table \ref{tab:baseline} shows a comparison of different feature combinations using mean overlap precision (OP) and area-under-the-curve (AUC). OP is defined as the percentage of frames in a video where the intersection-over-union overlap exceeds a certain threshold. In the tables, we report the OP at the threshold of $0.5$, which corresponds to the PASCAL criterion. The AUC score is computed from the success plot, where OP is plotted over the range of thresholds $[0, 1]$. See \cite{OTB2015} for further details about the evaluation metrics.

The results show that using only HOG gives a mean OP score of $74.5\%$. Interestingly, adding a deep RGB feature layer (RGB(d)) improves the result by $6.2\%$, while adding a deep motion feature layer provides an improvement of $6.8\%$. The best result are obtained by combining all these three cues: HOG, RGB(d) and deep motion. This combination achieves an absolute gain of $9.6 \%$ in mean OP over using only HOG and obtains the best AUC score of $67.4\%$. Another interesting comparison is that the result of using HOG or a shallow RGB layer (RGB(s)) alone both provide approximately the same mean OP and AUC score as their combination. This indicates that HOG and RGB(s) do not provide significant complementary information. On the other hand, adding deep motion features to either of these representations significantly improves the results. From our results, it is apparent that adding deep motion features consistently increases the tracking performance. Lastly, our results clearly show that deep appearance and motion features are complementary and that the best results are obtained when fusing these two cues.
For the state-of-the-art comparisons described in section \ref{sec:otb} and \ref{sec:tpl}, we employ the most successful feature combination in our approach, combining HOG, RGB(d) and deep motion features. Out tracker obtains an average fps of 0.0659, using precomputed optical flow maps. For future work, we aim to investigate speeding up the online learning and optical flow computation.

 \begin{figure}[!t]
 	\centering\vspace{-4mm}
 	\newcommand{\wid}{0.23\textwidth}
 	\subfloat[OTB-2015\label{fig:sota_otb}]{\includegraphics[trim={5 5 5 5},width = \wid]{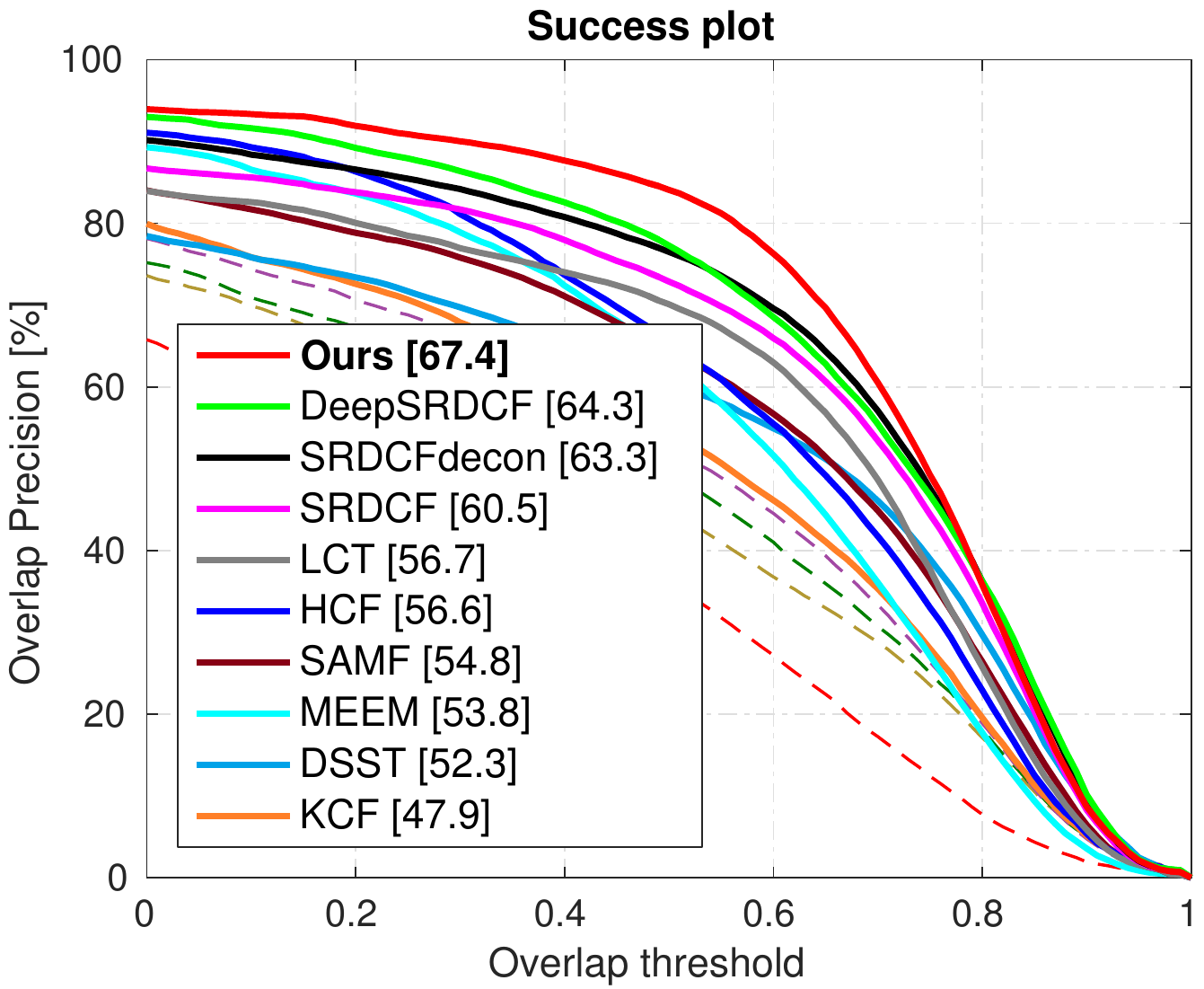}}\hspace{0.1cm}
 	\subfloat[Temple-Color\label{fig:sota_tpl}]{\includegraphics[trim={5 5 5 5},width = \wid]{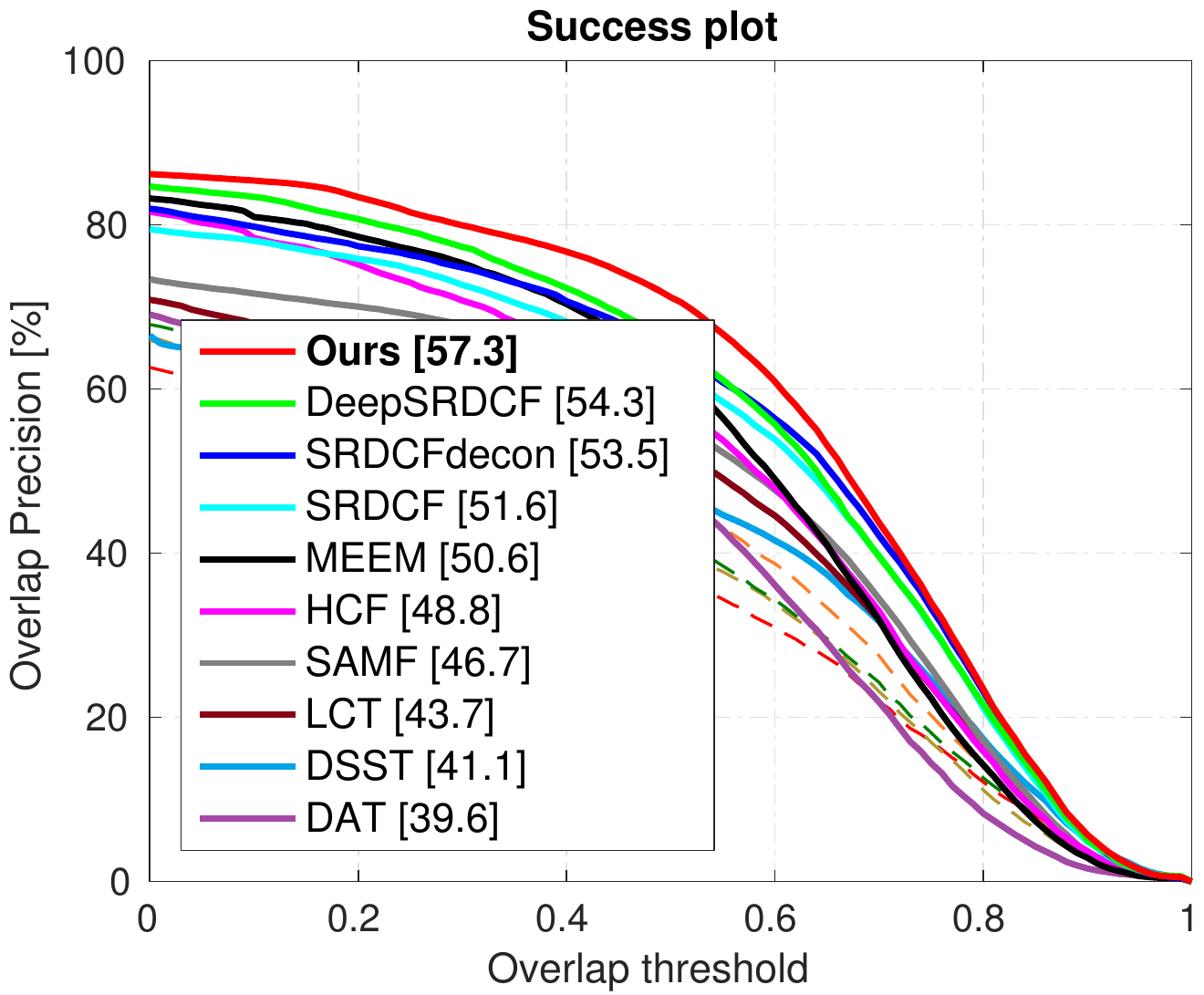}}\vspace{0mm}
 	\caption{Success plots showing a comparison of our approach with state-of-the-art methods on the OTB-2015 (a) and Temple-Color (b) datasets. For clarity, only the top 10 trackers are shown. Our proposed method provides significant improvements on both these datasets.}\vspace{-4mm}
 	\label{fig:OTB_tpl}
 \end{figure}

\subsection{OTB-2015 Dataset} \label{sec:otb}
We validate our approach in a comprehensive comparison with 12 state-of-the-art trackers: Struck \cite{Torr11b}, CFLB \cite{GaloogahiCVPR2015}, LCT \cite{LTC_CVPR15}, ACT \cite{DanelljanCVPR14}, KCF \cite{Henriques14}, DSST \cite{DanelljanBMVC14}, SAMF \cite{Li2014}, DAT \cite{possegger15a}, MEEM \cite{MEEM2014}, HCF \cite{HCF_ICCV15}, SRDCF \cite{DanelljanICCV2015} and SRDCFdecon \cite{DanelljanCVPR2016a}. We also compare with the DeepSRDCF \cite{DanelljanVOT2015}, that employs the shallow layer of a deep RGB network in the SRDCF. 

Table~\ref{tab:otb_tpl} (first row) presents a state-of-the-art comparison, in mean OP, on the OTB-2015 dataset. The HCF tracker employing an ensemble of deep RGB features obtains a mean OP score of $65.5\%$. The SRDCF tracker using hand-crafted appearance features achieves a mean OP score of $72.9\%$. The DeepSRDCF employs appearance based RGB features and obtains a mean OP score of $77.3\%$. Our approach that combines hand-crafted and deep appearance based features with deep motion features achieves state-of-the-art results on this datset with a mean OP of $84.1\%$. Figure~\ref{fig:sota_otb} presents the success plot for top-10 trackers on the OTB-2015 dataset. The area-under-the-curve (AUC) for each method is shown in the legend. The SRDCF obtains an AUC score of $60.5\%$. Among the existing methods, the DeepSRDCF achieves an AUC score of $64.3\%$. Our approach significantly outperforms the DeepSRDCF tracker by obtaining an AUC score of $67.4\%$.

\begin{table*}
	\centering
	\caption{Comparison with state-of-the-art methods on the VOT2015 dataset. The proposed method provides state-of-the-art accuracy and robustness.}
	\vspace{-2mm}
	\resizebox{0.8\textwidth}{!}{
		\begin{tabular}{lccccccccccc}
			\toprule
			&S3Tracker&RAJSSC&Struck&NSAMF&SC-EBT&sPST&LDP&SRDCF&EBT&DeepSRDCF&\textbf{Ours}\\\midrule
			Robustness&1.77&1.63&1.26&1.29&1.86&1.48&1.84&1.24&\textit{\textcolor{blue}{1.02}}&1.05&\textbf{\textcolor{red}{0.92}}\\
			Accuracy&0.52&\textit{\textcolor{blue}{0.57}}&0.47&0.53&0.55&0.55&0.51&0.56&0.47&0.56&\textbf{\textcolor{red}{0.58}}\\\bottomrule
		\end{tabular}}
		\vspace{-2mm}
		\label{tab:vot}
\end{table*}
\begin{figure*}[!t]
		\centering
		\newcommand{\wid}{0.25\textwidth}
		\newcommand{\name}{figures/sota_OTB100}
		\newcommand{\eval}{OPE}
		\includegraphics[width=\wid]{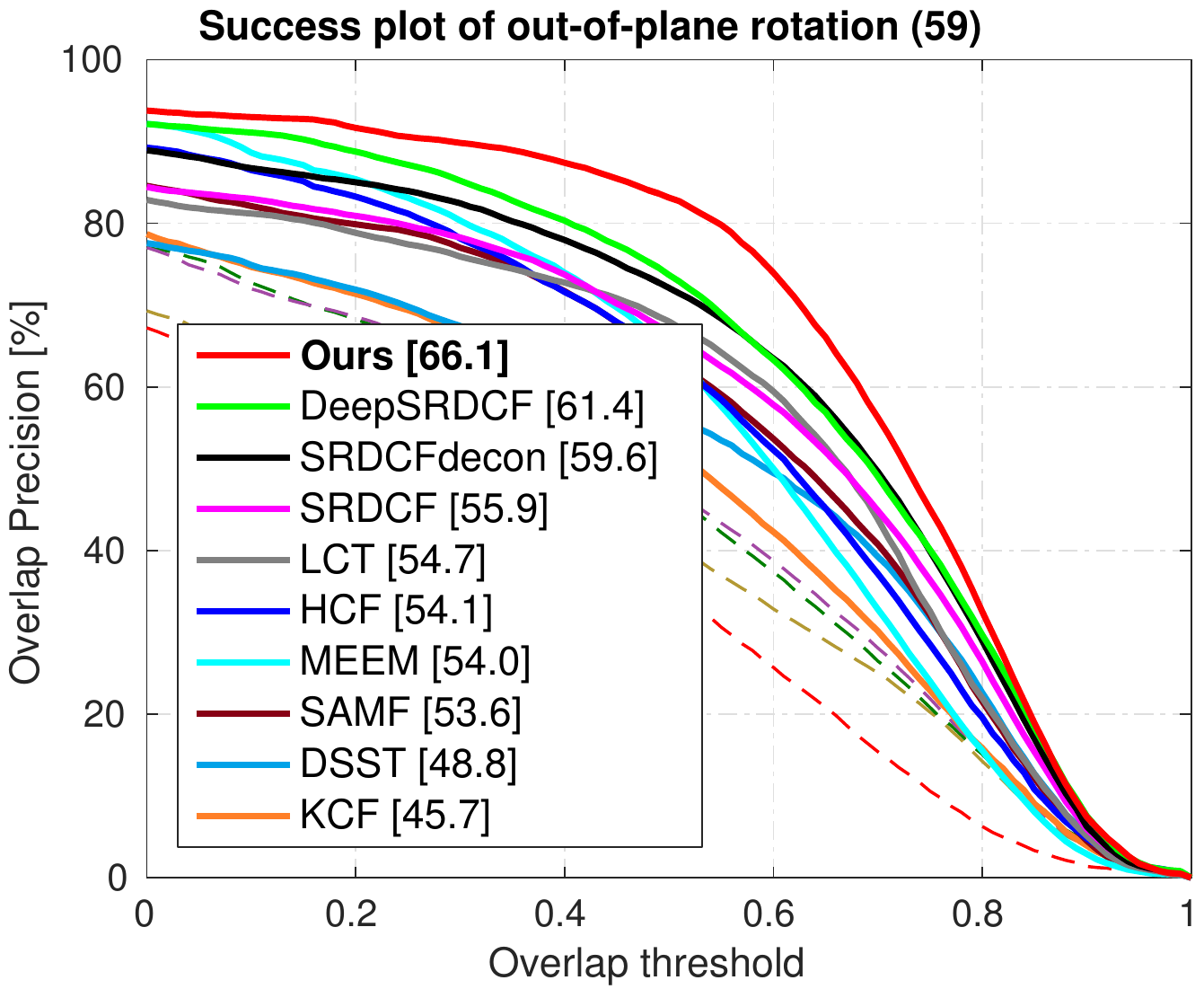}%
		\includegraphics[width=\wid]{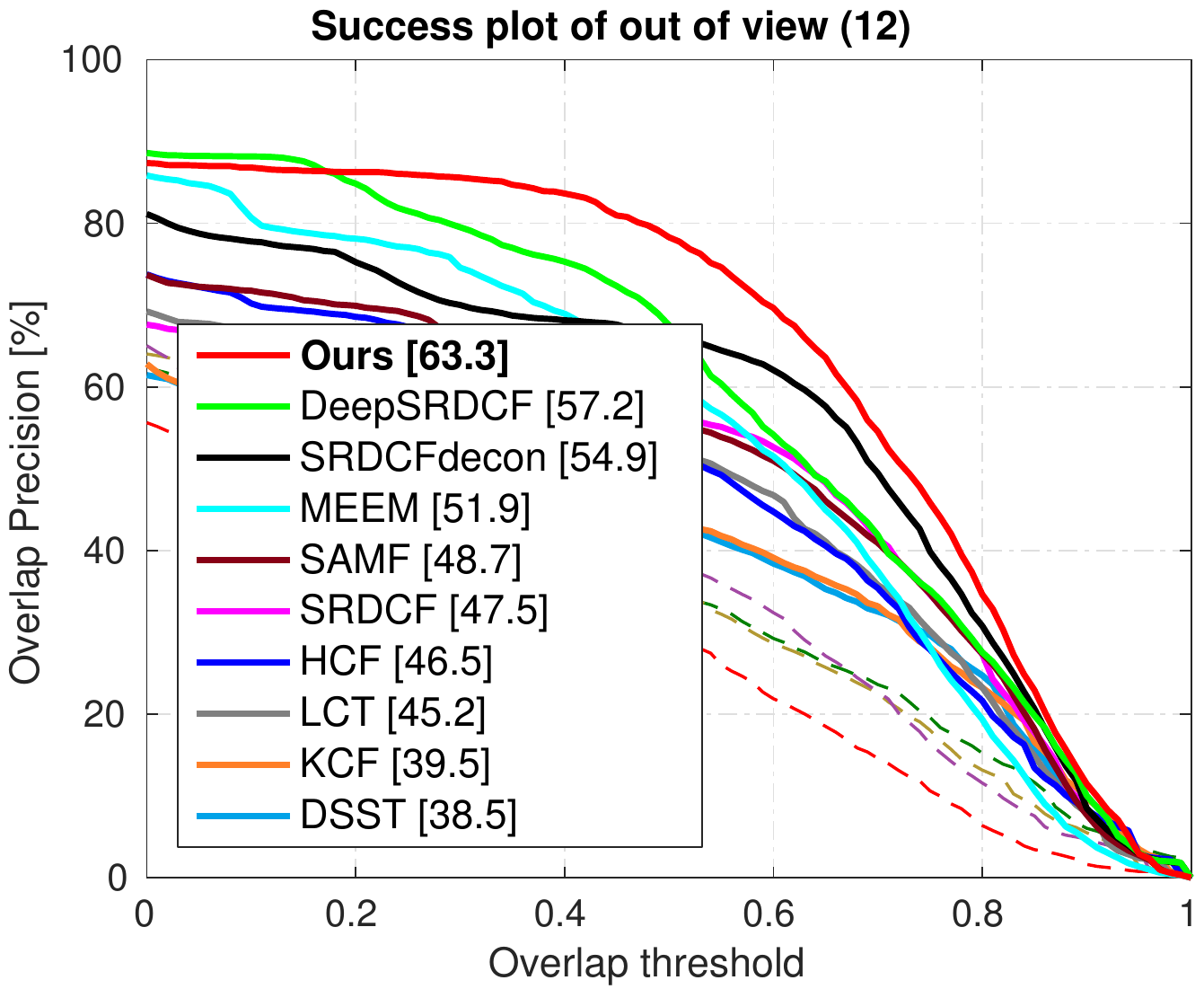}%
		\includegraphics[width=\wid]{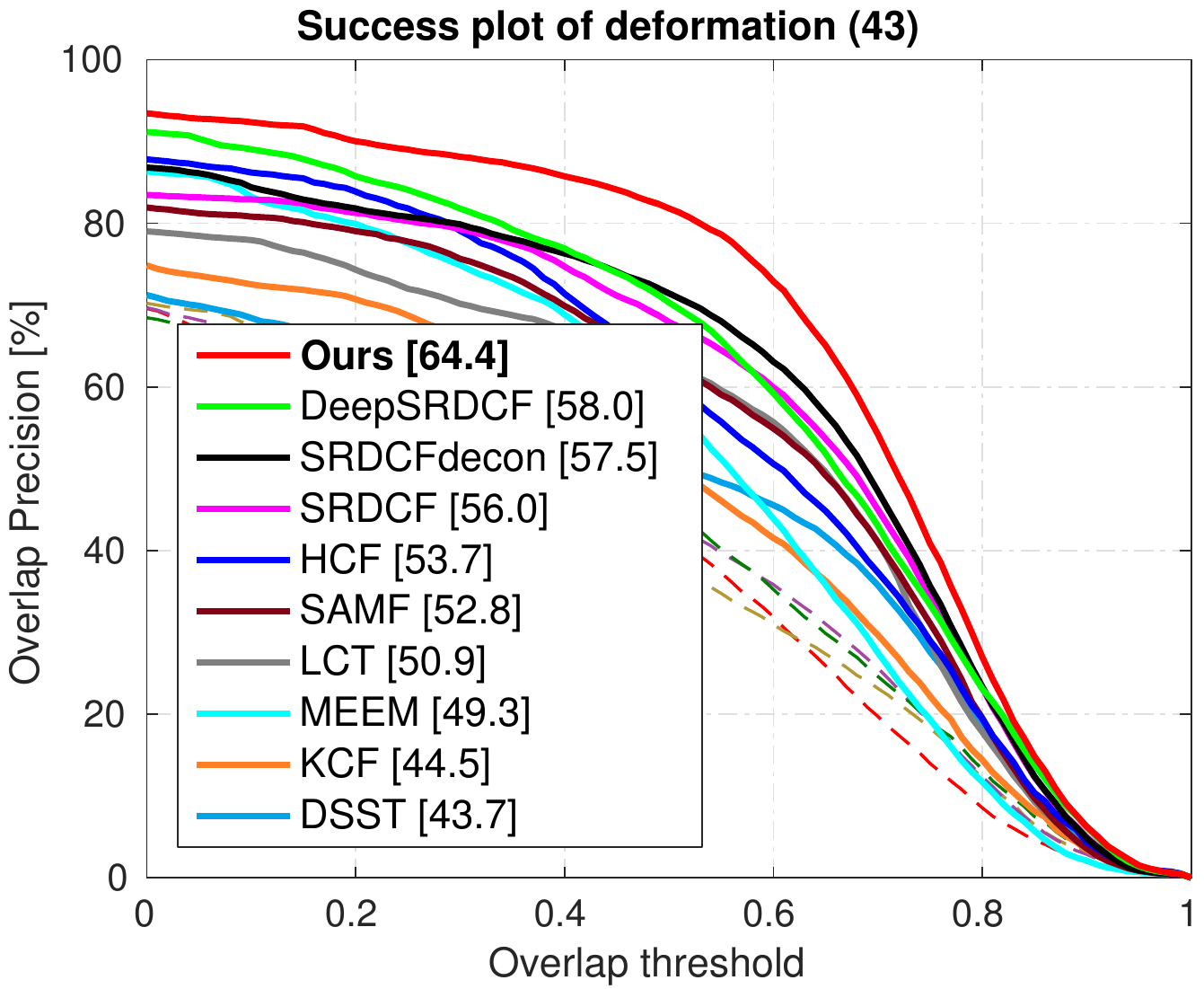}%
		\includegraphics[width=\wid]{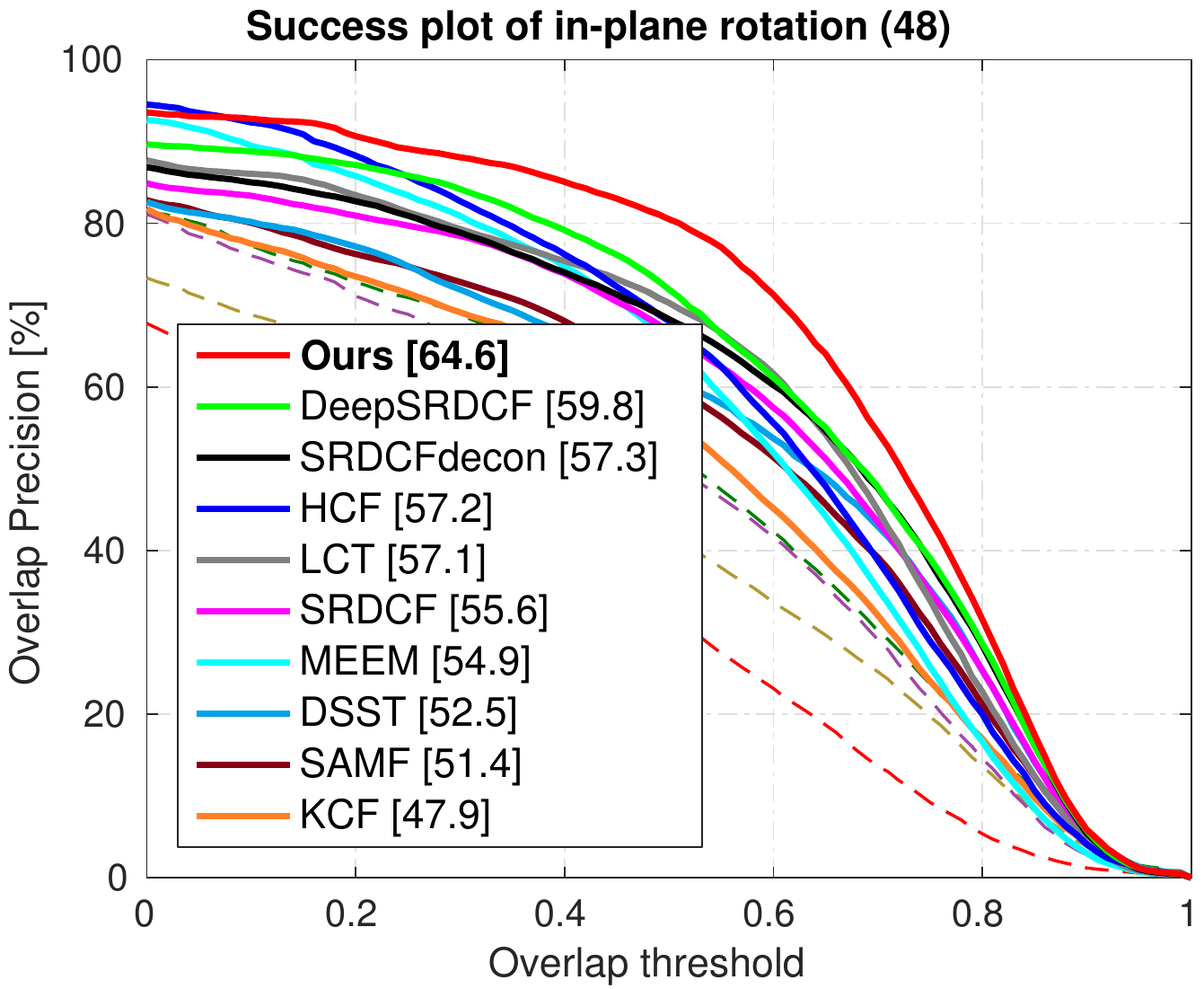}%
		\vspace{-3mm}
		\caption{Attribute-based analysis of our approach on the OTB-2015 dataset. Success plots are shown for 4 attributes. For clarity, we show the top 10 trackers in each plot. The title of each plot indicates the number of videos labeled with the respective attribute. Our approach provides consistent improvements compared to state-of-the-art methods on all 11 attributes.}
		\label{fig:attribute}
		\vspace{-4mm}
\end{figure*}

\noindent\textbf{Attribute-based Comparison:}
We perform an attribute-based analysis on the OTB-2015 dataset. Each video in the dataset is annotated by 11 different attributes: illumination variation, scale variation,  occlusion, deformation, motion blur, fast motion, in-plane and out-of-plane rotation,
out-of-view, background clutter and low resolution. Figure~\ref{fig:attribute} shows success plots for 4 attributes. Our approach provides consistent improvements on all 11 attributes. A significant improvement is achieved in these scenarios: deformation ($+6.4\%$), out of view ($+6.1\%$), and out-of-plane rotation ($+4.7\%$), compared to the best existing tracker.

\subsection{Temple-Color Dataset} \label{sec:tpl}
Next, we validate our proposed tracker on the challenging Temple-Color dataset \cite{TempleColor}. The dataset consists of 128 videos. Table~\ref{tab:otb_tpl} (second row) presents a state-of-the-art comparison in mean OP. The HCF tracker obtains a mean OP score of $58.2\%$. The SRDCF tracker with hand-crafted appearance features provides a mean OP score of $62.2\%$. The DeepSRDCF further improves the results and obtains a mean OP score of $65.4\%$. Our approach obtains state-of-the-art results on this datset with a mean OP of $71.2\%$. A significant gain of $5.8\%$ in mean OP is obtained over the DeepSRDCF tracker. Figure~\ref{fig:sota_tpl} presents the success plot for top-10 trackers on the Temple-Color dataset. The area-under-the-curve (AUC) for each tracker is shown in the legend of the plot. The SRDCF obtains an AUC score of $51.6\%$. Among the existing methods, the DeepSRDCF provides the best results and achieves an AUC score of $54.3\%$. Our approach obtains state-of-the-art results by significantly outperforming the DeepSRDCF tracker with a gain of $3\%$.

\subsection{VOT2015 Dataset}
Table~\ref{tab:vot} presents a state-of-the-art comparison on the VOT2015 dataset \cite{VOT2015} in comparison to the top 10 participants in the challenge according to the VOT2016 rules (see \url{http://votchallenge.net}). The dataset consists of 60 challenging videos compiled from a set of more than 300 videos. Here, the performance is measured in terms of accuracy (overlap with the ground-truth) and robustness (failure rate). The proposed method yields superior accuracy compared to the previously most accurate method (RAJSSC) and superior robustness compared to the previously most robust method (EBT). Compared to previous SRDCF-based methods, both accuracy and robustness are significantly improved.

\section{Conclusions}
We investigate the impact of deep motion features in a DCF-based tracking framework. Existing approaches are limited to using either hand-crafted or deep appearance based features. We show that deep motion features provide complementary information to appearance cue and their combination leads to significantly improved tracking performance. Experiments are performed on three challenging benchmark tracking datasets: OTB-2015 with 100 videos, Temple-Color with 128 videos, and VOT2015 with 60 videos. Our results clearly demonstrate that fusion of hand-crafted appearance, deep appearance and deep motion features leads to state-of-the-art performance on both datasets. Future work includes incorporating deep motion features into the recent C-COT \cite{DanelljanECCV2016} framework to enable joint fusion of the multi-resolution feature map.

\noindent\textbf{Acknowledgments}:
This work has been supported by SSF (CUAS), VR (EMC${}^2$), FFI (CCAIS), WASP, NSC and Nvidia.

{\small
\bibliographystyle{IEEEtran}
\bibliography{references}
}

\end{document}